\documentclass{article}
\usepackage{spconf}
\usepackage{amsmath,graphicx}
\graphicspath{{./img/}}

\usepackage{color}
\usepackage{amsmath}

\newcommand{\tabincell}[2]{\begin{tabular}{@{}#1@{}}#2\end{tabular}}

\title{A New Registration Approach for
Dynamic Analysis of Calcium Signals in Organs}
\name{Peixian Liang$^{1\star}$, Jianxu Chen$^{1\star}$, Pavel A. Brodskiy$^{2}$, Qinfeng Wu$^{2}$}
\secondlinename{Yejia Zhang$^{3}$, Yizhe Zhang$^{1}$, Lin Yang$^{1}$, Jeremiah J. Zartman$^{2}$ and Danny Z. Chen$^{1}$
\thanks{$\star$ These authors contribute equally to this work.}}

\address{$^{1}$ Department of Computer Science and Engineering, University of Notre Dame, USA \\
   $^{2}$Department of Chemical and Biomolecular Engineering, University of Notre Dame, USA\\
   $^{3}$Department of Electrical and Computer Engineering, University of California San Diego, USA}

\begin{document}
\ninept
\maketitle

\begin{abstract}
Wing disc pouches of fruit flies are a powerful genetic model for studying physiological intercellular calcium ($Ca^{2+}$) signals for dynamic analysis of cell signaling in organ development and disease studies. 
A key to analyzing spatial-temporal patterns of $Ca^{2+}$ signal waves is to accurately align the pouches across image sequences. However, pouches in different image frames may exhibit extensive intensity oscillations due to $Ca^{2+}$ signaling dynamics, and commonly used multimodal non-rigid registration methods may fail to achieve satisfactory results. In this paper, we develop a new two-phase non-rigid registration approach to register pouches in image sequences. First, we conduct segmentation of the region of interest. (i.e., pouches) using a deep neural network model. 
Second, we 
use a B-spline based registration to 
obtain an optimal transformation and align pouches across the image sequences. 
Evaluated using both synthetic data and real pouch data, our method considerably outperforms the state-of-the-art non-rigid registration methods. 

\end{abstract}

\section{Introduction}

$Ca^{2+}$ is a ubiquitous second messenger in organisms \cite{Wu-Zartman-2017}. Quantitatively analyzing spatial-temporal patterns of intercellular calcium ($Ca^{2+}$) signaling in tissues is important for understanding biological functions. 
Wing disc pouches of fruit flies are a commonly used genetic model system of organ development and have recently been used to study the decoding of $Ca^{2+}$ signaling in epithelial tissues \cite{Wu-Zartman-2017, ardiel2017visualizing}. 
However, wing discs can undergo considerable movements and deformations during live imaging experiments. Therefore, an effective automatic image registration approach is needed to align pouches across image sequences.

Registering tissues that undergo deformations in time-lapse image sequences is a challenging problem. For example, experimental image data of wing disc pouches
at different time points are moving or deforming due to a general feature of tissue growth, morphogenesis, and due to general movement during the live imaging process. 
Furthermore, a time-lapse movie can contain many frames, and a number of movies are needed to obtain reliable measurements of $Ca^{2+}$ activity due to the noisy and stochastic nature of the signals, which make the processing more complicated and costly. 
Common approaches suffer considerably due to cumbersome intensity distortions of tissues caused by $Ca^{2+}$ oscillations.
A method for minimizing the error residual between the local phase-coherence representations of two images was proposed to deal with non-homogeneity in images \cite{wong2009robust}, which relies heavily on structural information. But, in our problem, the intensity inside the pouches may change a lot, thus causing such methods to fail. A Markov-Gibbs random field model with pairwise interaction was used to learn prior appearance of a given prototype, which makes it possible to align complex images \cite{el2006image}. Incorporation of spatial and geometric information was proposed to address the limitations of the static local intensity relationship \cite{woo2015multimodal}. But,  the computational complexity of these methods is high.
In our problem, a single time-lapse movie may have hundreds of frames, and hundreds of movies are analyzed. Hence, these methods do not work well for our problem. In general, known methods do not address well the kind of complex intensity distortion in our images with a reasonable computation cost.

\begin{figure}
\centering
\includegraphics[width = \linewidth]{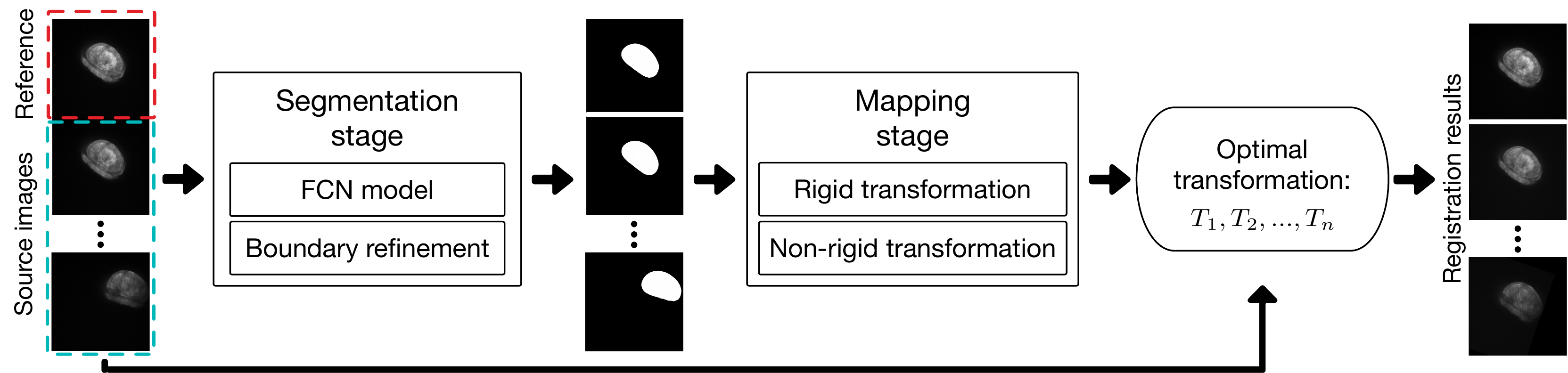}
\caption{An overview of our proposed registration approach. }
\label{fig:overview}
\end{figure}

To address the difficulty of spatial intensity distortion along with various ROI deformations and movements, we propose a new non-rigid registration approach to effectively align pouches at different time points in image sequences. Our approach consists of two main stages: (1) a segmentation stage and (2) a mapping stage (see Fig.~\ref{fig:overview}). The segmentation stage is based on a deep neural network (FCN). Because the accuracy of the ROI boundary is a key factor influencing the registration results, we apply a graph search algorithm to refine the segmented pouch boundaries. The mapping stage uses the segmentation results from the first stage to characterize an optimal transformation. We first apply a rigid transformation to modulate the movement of pouches, and then design an object pre-detection B-spline based non-rigid algorithm to produce the final mapping.


\section{Methodology}
 
In this section, we present a complete pipeline, which takes time-lapse image sequences of pouches as input and produces the registered sequences (see Fig.~\ref{fig:overview}). We first discuss the segmentation stage (i.e., FCN model and boundary refinement), and then the mapping stage (i.e., rigid transformation and non-rigid transformation).

\subsection{Segmentation Stage}

Our registration approach is based on accurate pouch segmentation. Pouches in our images are commonly surrounded and touched by extra tissues with similar intensity and texture, and the separation boundary between pouches and extra tissues is usually of poor visibility. 
Meanwhile, the noise induced by the live imaging process 
makes the segmentation task more challenging. Thus, it is important for our segmentation algorithm to leverage the morphological and topological contexts, in order to correctly segment the shape of each actual pouch, especially its boundary, from the noisy background. For this, we employ an FCN model to exploit the semantic context for accurate segmentation, and a graph search algorithm to further refine the boundaries.


\noindent
\textbf{FCN module}\hspace{5pt} Recently, deep learning methods have emerged as powerful image segmentation tools. Fully convolutional networks (FCN) are widely used in general semantic segmentation 
and biomedical image segmentation \cite{nips,ronneberger2015u}. 

\begin{figure}
\centering
\includegraphics[width = \linewidth]{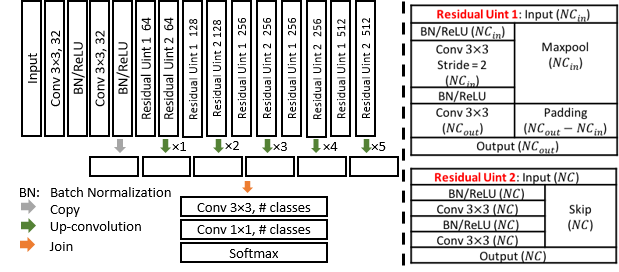}
\caption{The structure of our FCN model.}
\label{DNN}
\end{figure}

It is worth mentioning that in our images, the separation boundary between a pouch and other tissues is usually quite subtle (as thin as 3 to 5 pixels wide) and obscure, while the whole contextual region (including both the pouch and extra tissues) can be of a relatively large scale (more than $200\times200$ pixels). Therefore, the FCN model must fully exploit both the fine details and a very large context. For this purpose, we carefully design the FCN architecture following the model in \cite{yang2017suggestive} to leverage a large receptive field without sacrificing model simplicity and neglecting fine details. The exact structure of our FCN model is depicted in Fig.~\ref{DNN}. 

\noindent
\textbf{Boundary refinement}\hspace{10pt} We observed that the boundaries of the output pouches from our FCN model may be fuzzy or of irregular shape in difficult cases (see Fig.~\ref{segmentation}(F)). To improve the boundary accuracy of the segmented pouches, we first apply a Gaussian smooth filter to reduce the influence of intensity variations inside the pouches and then employ a graph search algorithm with the node weights being the negatives of the gradients to further refine the shape boundaries. This allows the subsequent mapping stage to be built upon more accurate segmentation and produce better registration results. Details of this process are omitted due to the page limit.

\subsection{Mapping Stage}

The goal of the registration process is: For every point in the source image, obtain an optimal corresponding point in the reference image. A key observation is that the intensity profile of the same pouch may incur substantial changes in different frames of an image sequence, due to undergoing $Ca^{2+}$ signal waves. Hence, intensity is not a reliable cue for finding optimal correspondence between points in different frames of a sequence. Here, we utilize the results from the segmentation stage. 

\noindent
\textbf{Rigid transformation}\hspace{10pt} Since there are lots of movements (i.e., rotation and translation) of pouches in image sequences, we first compute an optimal rigid transformation to reduce their influence. This optimization step uses a regular-step gradient descent algorithm. Note that here, local optimum traps could be an issue in practice. Specifically, since the pouches are often of oval shapes, a local optimum may yield incorrect results where the object is aligned with the opposite orientation.
To resolve this issue, we always initialize the optimization process using the optimum parameters computed from the preceding frame in the sequence, since the pouch movement in consecutive frames is usually not very big.

\noindent
\textbf{Non-rigid transformation}\hspace{5pt}
The non-rigid registration seeks an optimal transformation $T$: $(x,y,t)\rightarrow (x',y',t_{0})$, which maps any points in the source image at time $t$ to time $t_{0}$ (i.e., in the reference image). We use the free-form deformation (FFD) transformation model, based on B-splines \cite{lee1996image}, to find an optimal transformation. 

\begin{figure}[!b]
\centering\includegraphics[width = 0.6\linewidth]{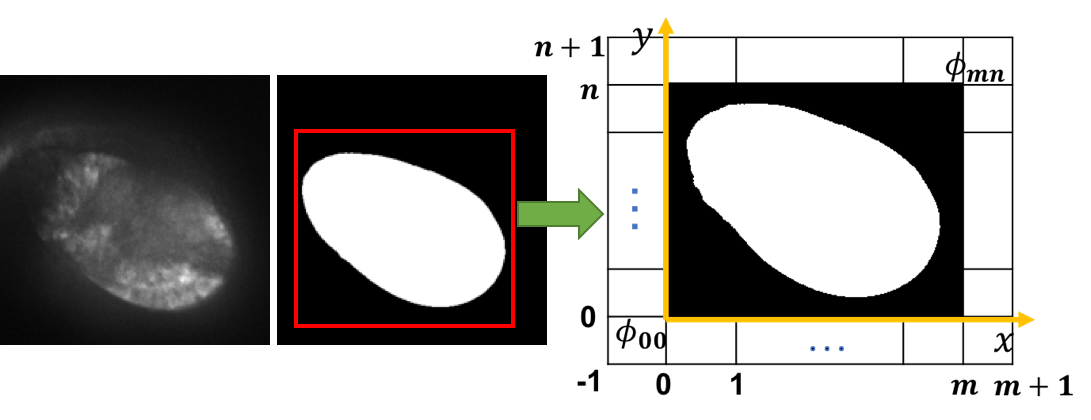}
\caption{Configuration of lattice grid $\Phi$ for non-rigid transformation.}
\label{space}
\end{figure}

For our problem, not too many control points are needed outside a pouch, because we need to focus only on the ROI. Thus, detecting ROIs first can save computation time. Pouches will be near the same position in the frames after the rigid registration, making it possible to do non-rigid transformation only around the ROI area. Based on this observation, we can crop an area in the first frame, and apply a lattice $\Phi$ to this area in the following changing frames. We define the lattice $\Phi$ as an $(m + 3) \times (n+3)$ grid in the domain $\Omega$ (see Fig.~\ref{space}). 

We define the registration model as follows. Let $\Omega = \{(x,y) \ | $ $ X_{l}\le x<X_{r}, Y_{l}\le y<Y_{r}\}$ be a rectangular domain in the $xy$ plane, where the $X$ and $Y$ values specify the boundary of the detection area. To approximate the intensity of scattered points, $I(x,y)$, in a pouch, we formulate a function $f$ as a uniform cubic B-spline function:
$f(x,y) = \sum\limits_{j=0}^3\sum\limits_{i=0}^3B_i(s)B_j(t)\Phi_{(i+k,j+l)}$, 
where $s=x-\lfloor x \rfloor, t=y - \lfloor y \rfloor,k=\lfloor x \rfloor - 1$, and $l=\lfloor y\rfloor-1$. In addition, $B_{i}$ represents the $i$-th basis function of the cubic B-spline:
$B_{0}(t)=(1-t)^{3}/6, B_{1}(t)=(3t^{3}-6t^{2}+4)/6, B_{2}(t)=(-3t^{3}-6t^{2}+4)/6$, and
$B_{3}(t)=(t^{3})/6.$

Since the resolution of the control points determines the non-rigid degree and has a big impact on the computation time, a higher resolution of control points gives more freedom to do deformation while also increasing the computation time. To optimize this trade-off, we use a multi-level mesh grid approach \cite{rueckert1999comparison} to devise a computationally efficient algorithm. Let $\Phi_{1},\Phi_{2},\ldots,\Phi_{g}$ denote a hierarchy of meshes of control points with increasing resolutions and $T_{1}, T_{2}, \ldots,T_{g}$ be the deformation functions with respect to each mesh. We first apply a coarse deformation, and then refine the deformation gradually. The size of the mesh grid is increased by a factor of 2 with the same spacing, so that the raw image is down-sampled to the corresponding size at different levels. The final deformation is the composition of these functions:
$T(\Omega) = T_{g}(...T_{2}(T_{1}(\Omega))...).$

To obtain an optimal transformation $\Phi$, we minimize the following energy function:
$E = F_{s} + QS$,
where the first term is the similarity measure and the second term is for regularization. $Q$ is the curvature penalization
and $S$ is the displacement of control points.
The similarity measure we use is the sum of squared distance (SSD):
$SSD = \frac{1}{N}\sum{(A-B^{T})^{2}}$, where $A$ is the reference image intensity function, $B^{T}$ is the intensity function of the transformed image of a source image $B$ under the current transformation $T$ and $N$ is the number of pixels.

This process iteratively updates the transformation parameters, $T$, using a gradient descent algorithm. 
When a local optimum of the cost function is smaller than $\epsilon$ or the number of iterations is bigger than $N_{max}$, the algorithm will terminate.

\section{Experiments and Evaluations} 

We evaluate our registration approach from two perspectives. First, we evaluate the accuracy of the segmentation method,  
because the segmentation accuracy is crucial to our overall approach. 
Second, we conduct experiments using both synthetic data and real biological data to assess the registration performance of different approaches. 

\subsection{Segmentation Evaluations} 

\begin{figure}[!b]
\centering
\includegraphics[width = 0.7\linewidth]{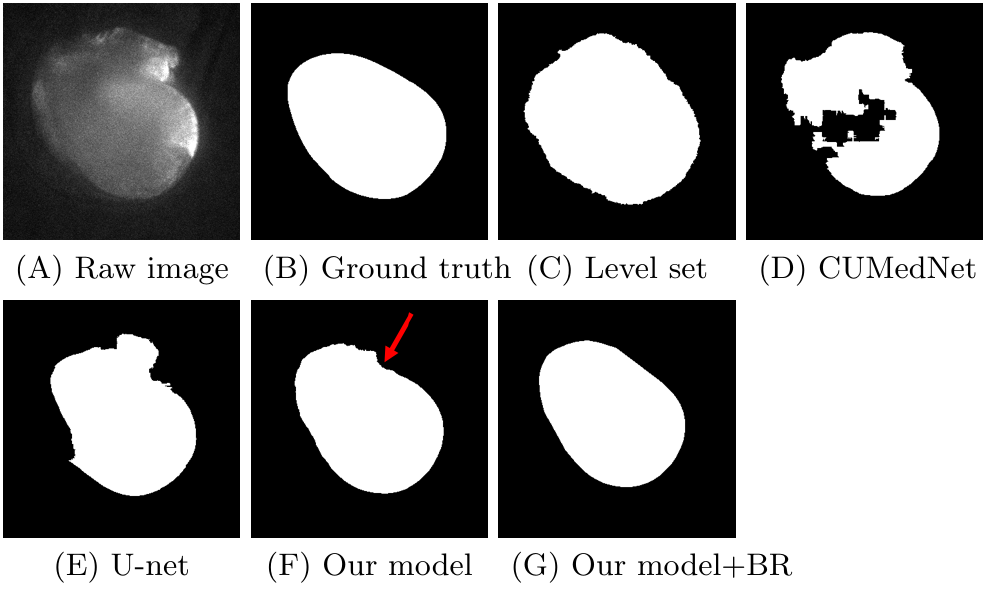}
\caption{Segmentation results of different methods on a somewhat difficult pouch image (BR = Boundary Refinement). The red arrow indicates a location that demonstrates the effect of BR.}
\label{segmentation}
\end{figure}

We conduct experiments on 100 images of 512$\times$512 pixels selected from 10 randomly chosen control videos. 
Our method is compared with both traditional method (e.g., level set \cite{li2008minimization}) and state-of-the-art FCN models (i.e., U-Net \cite{ronneberger2015u} and CUMedNet \cite{chen2016deep}).
The FCN models 
are trained using the Adam optimizer \cite{adam} with a learning rate of 0.0005. Data augmentation (random rotation and flip) is used during training. 
We use the mean IU (intersection over union) and F1 as the metrics.
Table~\ref{table:best} shows the quantitative results of different methods, and Fig.~\ref{segmentation} shows some segmentation examples of pouches using different methods. 
It is evident that our FCN model works better in segmenting difficult ROIs, and our boundary refinement can help obtain accurate ROI boundaries.

\begin{table}  
\caption{Comparison results of segmentation.}
\centering
\begin{tabular}{| l | l | l |l|}
\hline
& Mean IU & F1 score \\ \hline
Level set & 0.8235 &  0.8294  \\\hline
CUMedNet & 0.9394 & 0.9454 \\\hline
U-Net & 0.9479 & 0.9542  \\\hline
Our FCN & 0.9586 & 0.9643 \\\hline

Our FCN + BR & 0.9617 & 0.9682 \\\hline
\end{tabular}
\label{table:best}
\end{table}

\subsection{Registration Evaluations}

 We compare the registration performance with the Demon algorithm \cite{kroon2009mri} and a B-spline method based on Residual Complexity (RC) \cite{myronenko2009image}, which are two state-of-the-art non-rigid registration methods for images with spatially-varying intensity. 

\begin{figure}[t]
\centering
\includegraphics[width = 0.7\linewidth]{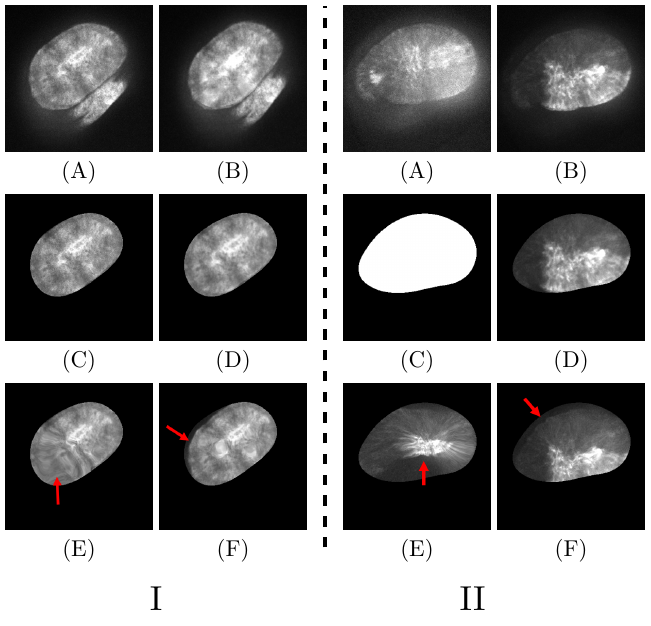}
\caption{Registration results. I: Synthetic data. (A) A reference image; (B) A source image; (C) The expected registration result; (D) Our method; (E) RC; (F) Demon. \uppercase\expandafter{\romannumeral2}: Real pouches. (C) Intermediate segmentation; (D) Our method; (E) RC; (F) Demon. The red arrows point at some areas with unsatisfactory registration.
}
\label{result}
\end{figure}

\noindent
\textbf{Synthetic data}\hspace{10pt}For synthetic data, we choose 8 pouch images as reference images.
Specifically, for each reference image $R$, we 
generate 20 source images
by adding geometric distortion $G_{T}$, and intensity distortion $I_{T}$, to simulate an image sequence with undergoing movements and $Ca^{2+}$ signaling (i.e., $R  \stackrel{G_{T}}{\longrightarrow} S_{i}^{1}
\stackrel{I_{T}}{\longrightarrow} S_{i}^{2}, i = 1,2,\ldots,20)$. For geometric distortions, we apply an elastic spline deformation to perturb the points according to the grid deformation and a rigid transformation. 
For intensity distortion, we first add Gaussian noise in random disk-shaped regions within the pouch to simulate the calcium signal waves and then rescale the intensity to [0, 1].
Our target is to find the optimal transformation $T_i$ from every source image $S_{i}^{2}$ to the corresponding reference image (i.e., $S_{i}^2\stackrel{T_i}{\longrightarrow}R)$. To quantify the performance, we compare the intensity root mean square error (RMSE) between the reference image $R$ and the clean registered images $C_{i}$, which are obtained by applying $T_i$ to the geometrically distorted images without intensity distortion (i.e., $S_{i}^{1}\stackrel{T_i}{\longrightarrow}C_{i})$. The idea is to evaluate whether the registration algorithm is able to find an optimal geometric transformation without damaging the texture. 
Fig.~\ref{result}.\uppercase\expandafter{\romannumeral1} shows some visual results of different methods, and Table~\ref{tab:setup} gives quantitative results. The results of our approach are considerably more accurate.

\begin{table}[ht!]

\centering
\caption{Registration results of synthetic data.}\label{tab:setup}
\begin{tabular}{|c|c|c|c|} \hline
 \tabincell{c}{Movie\\No.} & \tabincell{c}{Our method\\ RMSE (pixel)} & \tabincell{c}{RC\\ RMSE (pixel)} & \tabincell{c}{Demon\\ RMSE (pixel)}
 \\ \hline
$1$ & $0.097\pm0.019$  & $0.137\pm0.039$ & $0.150\pm0.023$\\ \hline
$2$  &$0.097\pm0.021$  & $0.133\pm0.025$ & $0.184\pm0.043$\\ \hline
$3$ & $0.091\pm0.012$ & $0.137\pm0.033$& $0.162\pm0.021$ \\ \hline
$4$ & $0.098\pm0.009$ & $0.119\pm0.024$ & $0.128\pm0.018$\\ \hline
$5$ & $0.086\pm0.011$ & $0.094\pm0.020$ &$0.124\pm0.010$\\ \hline
$6$ & $0.088\pm0.013$ & $0.132\pm0.059$ & $0.149\pm0.015$\\ \hline
$7$ & $0.095\pm0.018$ & $0.096\pm0.013$ &$0.133\pm0.021$\\ \hline
$8$ & $0.099\pm0.019$ & $0.109\pm0.019$ &$0.175\pm0.034$\\ \hline
$Average$ & $0.094\pm0.015$ & $0.120\pm0.029$ &$0.151\pm0.023$\\ \hline
\end{tabular}
\end{table}

\begin{table}[ht!]
\centering
\caption{Registration results of real pouch data.}\label{realData}
\begin{tabular}{|c|c|c|c|} \hline
 \tabincell{c}{Movie \\No. }& \tabincell{c}{Our method\\HD (pixel)} & \tabincell{c}{RC\\HD (pixel)} & \tabincell{c}{Demon\\HD (pixel)}
 \\ \hline
$1$ & $5.199\pm0.484$  & $5.490\pm0.876$ & $5.633\pm0.517$\\ \hline
$2$  &$5.556\pm1.250$  & $6.189\pm0.833$ & $6.819\pm1.193$ \\\hline
$3$ & $5.568\pm0.702$ & $6.403\pm0.812$& $6.164\pm0.514$\\ \hline
$4$ & $4.342\pm1.347$ & $6.160\pm1.311$ & $6.761\pm0.787$\\ \hline
$5$ & $4.622\pm0.684$ & $5.461\pm0.824$ &$5.072\pm0.330$\\ \hline
$6$ & $4.913\pm0.527$ & $5.328\pm1.394$ & $6.696\pm0.703$\\ \hline
$7$ & $6.882\pm1.162$ & $7.413\pm0.574$ &$7.192\pm0.836$\\ \hline
$8$ & $5.344\pm0.443$ & $6.173\pm0.805$ &$5.845\pm0.576$\\ \hline
$Average$ & $5.303\pm0.825$ & $6.077\pm0.929$ &$6.273\pm0.682$\\ \hline
\end{tabular}
\end{table}

\noindent
\textbf{Wing disc pouch data}\hspace{8pt}
 We randomly choose 8 movies from 150 control videos. 
 In each movie, we take the first frame as the reference image and all the other frames as source images. 
 To validate the registration results, we apply transformation $T$ to 
 the annotation of source images to obtain the registered annotation boundary and compare the results using the Hausdorff distance (HD) error metric.  
 Table~\ref{realData} gives the quantitative results. Our approach achieves accurate boundary shapes. Also, our method obtains clear texture inside ROIs, as shown by the examples in Fig.~\ref{result}.\uppercase\expandafter{\romannumeral2.}



\section{Discussions and Conclusions}
In this paper, we propose a new two-stage non-rigid image registration approach and apply it to analyze live imaging data of wing disc pouches 
for $Ca^{2+}$ signaling study. 
Comparing to the state-of-the-art non-rigid methods for biomedical image registration, our approach achieves higher accuracy in aligning images with non-negligible texture distortions. Our approach lays a foundation for quantitative analysis of pouch image sequences in whole-organ studies of $Ca^{2+}$ signaling related diseases.
Our approach may be extended to solving other biomedical image registration problems, especially when the intensity profiles and texture patterns of the target objects incur significant changes. The mapping stage of our approach is application-dependent, while our segmentation method is general and can be applied to many problems by modifying only the graph search based boundary refinement procedure. 

\section{Acknowledgment}

This research was supported in part by the Nanoelectronics Research Corporation (NERC), a wholly-owned subsidiary of the Semiconductor Research Corporation (SRC), through Extremely Energy Efficient Collective Electronics (EXCEL), an SRC-NRI Nanoelectronics Research Initiative under Research Task ID 2698.005, and by NSF grants CCF-1640081, CCF-1217906, CNS-1629914, CCF-1617735, CBET-1403887, and CBET-1553826, NIH R35GM124935, 
Harper Cancer Research Institute Research like a Champion awards, Walther Cancer Foundation Interdisciplinary Interface Training Project, the Notre Dame Advanced Diagnostics and Therapeutics Berry Fellowship, 
the Notre Dame Integrated Imaging Facility, the Bloomington Stock Center for fly stocks. 



%

\bibliographystyle{ieeetr}
\bibliography{paper}

\begin{thebibliography}{10}

\bibitem{Wu-Zartman-2017}
Q.~Wu, P.~Brodskiy, C.~Narciso, M.~Levis, J.~Chen, P.~Liang,
  N.~Arredondo-Walsh, D.~Z. Chen, and J.~J. Zartman, ``Intercellular calcium
  waves are controlled by morphogen signaling during organ development,'' {\em
  bioRxiv}, no.~104745, 2017.

\bibitem{ardiel2017visualizing}
E.~L. Ardiel, A.~Kumar, J.~Marbach, R.~Christensen, R.~Gupta, W.~Duncan, J.~S.
  Daniels, N.~Stuurman, D.~Col{\'o}n-Ramos, and H.~Shroff, ``Visualizing
  calcium flux in freely moving nematode embryos,'' {\em Biophysical Journal},
  vol.~112, no.~9, pp.~1975--1983, 2017.

\bibitem{wong2009robust}
A.~Wong and J.~Orchard, ``Robust multimodal registration using local
  phase-coherence representations,'' {\em Journal of Signal Processing
  Systems}, vol.~54, no.~1-3, p.~89, 2009.

\bibitem{el2006image}
A.~El-Baz, A.~Farag, G.~Gimel'farb, and A.~E. Abdel-Hakim, ``Image alignment
  using learning prior appearance model,'' in {\em ICIP}, pp.~341--344, 2006.

\bibitem{woo2015multimodal}
J.~Woo, M.~Stone, and J.~L. Prince, ``Multimodal registration via mutual
  information incorporating geometric and spatial context,'' {\em IEEE
  Transactions on Image Processing}, vol.~24, no.~2, pp.~757--769, 2015.

\bibitem{nips}
J.~Chen, L.~Yang, Y.~Zhang, M.~Alber, and D.~Z. Chen, ``Combining fully
  convolutional and recurrent neural networks for {3D} biomedical image
  segmentation,'' in {\em NIPS}, pp.~3036--3044, 2016.

\bibitem{ronneberger2015u}
O.~Ronneberger, P.~Fischer, and T.~Brox, ``{U-Net}: Convolutional networks for
  biomedical image segmentation,'' in {\em MICCAI}, pp.~234--241, 2015.

\bibitem{yang2017suggestive}
L.~Yang, Y.~Zhang, J.~Chen, S.~Zhang, and D.~Z. Chen, ``Suggestive annotation:
  A deep active learning framework for biomedical image segmentation,'' in {\em
  MICCAI}, pp.~399--407, 2017.

\bibitem{lee1996image}
S.~Lee, G.~Wolberg, K.-Y. Chwa, and S.~Y. Shin, ``Image metamorphosis with
  scattered feature constraints,'' {\em IEEE Transactions on Visualization and
  Computer Graphics}, vol.~2, no.~4, pp.~337--354, 1996.

\bibitem{rueckert1999comparison}
D.~Rueckert, L.~Sonoda, E.~Denton, S.~Rankin, C.~Hayes, M.~O. Leach, D.~Hill,
  and D.~J. Hawkes, ``Comparison and evaluation of rigid and non-rigid
  registration of breast {MR} images,'' in {\em SPIE}, vol.~3661, pp.~78--88,
  1999.

\bibitem{li2008minimization}
C.~Li, C.-Y. Kao, J.~C. Gore, and Z.~Ding, ``Minimization of region-scalable
  fitting energy for image segmentation,'' {\em IEEE Transactions on Image
  Processing}, vol.~17, no.~10, pp.~1940--1949, 2008.

\bibitem{chen2016deep}
H.~Chen, X.~Qi, J.-Z. Cheng, P.-A. Heng, {\em et~al.}, ``Deep contextual
  networks for neuronal structure segmentation.,'' in {\em AAAI},
  pp.~1167--1173, 2016.

\bibitem{adam}
D.~Kingma and J.~Ba, ``Adam: A method for stochastic optimization,'' {\em arXiv
  preprint arXiv:1412.6980}, 2014.

\bibitem{kroon2009mri}
D.-J. Kroon and C.~H. Slump, ``{MRI} modality transformation in demon
  registration,'' in {\em ISBI}, pp.~963--966, 2009.

\bibitem{myronenko2009image}
A.~Myronenko and X.~Song, ``Image registration by minimization of residual
  complexity,'' in {\em CVPR}, pp.~49--56, 2009.

\end{thebibliography}

\end{document}